# Proceedings of the Thirteenth International Workshop on Juris-informatics (JURISIN 2019)

*hosted by JSAI-isAI2019*

Workshop Co-Chairs

Makoto Nakamura
Satoshi Tojo

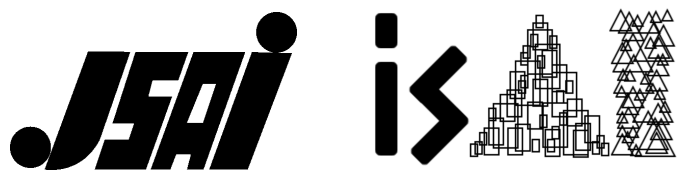

Raiosha Building, Keio University, Kanagawa, Japan,
November 11-12, 2019

# Measuring Patent Claim Generation by Span Relevancy

Jieh-Sheng Lee [*] [0000-0002-0990-6170] and Jieh Hsiang

Department of Computer Science and Information Engineering
National Taiwan University, Taiwan
{d04922013, hsiang}@csie.ntu.edu.tw

**Abstract.** Our long-term goal of patent claim generation is to realize "augmented inventing" for inventors by leveraging new Deep Learning techniques. We envision the possibility of building an "auto-complete" function for inventors to conceive better inventions in the era of artificial intelligence. In order to generate patent claims with reasonable quality, a fundamental question is how to measure the quality. We tackle the problem from the perspective of claim span relevancy as a proof of concept. Patent claim language was rarely explored in the NLP field. In this work, we propose a span-based approach and a generic framework to measure patent claim generation quantitatively. In order to study the effectiveness of patent claim generation, we define a metric to measure whether two consecutive spans in a generated patent claims are relevant. We treat such relevancy measurement as a span-pair classification problem, following the concept of natural language inference. Technically, the span-pair classifier is implemented by fine-tuning a pre-trained language model. The patent claim generation is implemented by fine-tuning the other pre-trained model. Specifically, we fine-tune a pre-trained Google BERT model to measure the patent claim spans generated by a fine-tuned OpenAI GPT-2 model. In this way, we re-use two of the state-of-the-art pre-trained models in the NLP field. Our result shows the effectiveness of the span-pair classifier after fine-tuning the pre-trained model. It further validates the quantitative metric of span relevancy in patent claim generation. Particularly, we found that the span relevancy ratio measured by BERT becomes lower when the diversity in GPT-2 text generation becomes higher.

---

# 1 Introduction

## 1.1 Patent Law & Deep Learning

Patents are granted to inventions that meet three basic legal requirements in general: utility, novelty, and nonobviousness. Utility is the requirement that an invention must have a useful function of some kind. Novelty is the requirement that an invention must be substantially different from everything that has been published or known before. Nonobviousness is the requirement that an invention cannot have been obvious to a person having ordinary skill in the art. From the perspective of NLP and Deep Learning, we hypothesize that the nonobviousness problem is a reinforcement learning problem between inventor/patent practitioner and patent examiner. The novelty problem is a search problem in essence. The utility problem is an NLI (Natural Language Inference) problem. In this paper, we focus on the utility perspective in the patent claim generation problem. Without a way to measure the likelihood of meeting the utility requirement, a patent claim generator may generate something novel and nonobvious but not useful at all.

## 1.2 Augmented Inventing

The ultimate goal of our research is an "augmented inventing" system. We envision an "auto-complete" use case in which, if an inventor is just contemplating and has no whole picture in mind yet, a function like patent claim generation may augment the inventor to conceive better inventions. For example, the interactive augmented-inventing system can suggest next words, phrases, claim spans or even new ideas based on user's input. Such active learning between human and machine may open a window for both qualitative and quantitative analysis on augmented inventing. By measuring how the inventor responds to the system, it is possible to collect human annotations for supervised learning. In order to facilitate supervised learning in the future, it is essential to generate patent claims with reasonable quality for inventors to appreciate. This paper is a step toward such a direction. We measure the quality by span relevancy and assume that a suitable range of span relevancy means a reasonable quality. It is noted that the relevancy measurement is implemented in an unsupervised fashion. By doing so, we have a chance to combine both unsupervised learning and supervised learning in the future.

## 1.3 A Span-based Approach

In the NLP field, language modeling is the task of predicting what word comes next. Instead of working on word level, we propose a span-based modeling approach to predict what text span may come next. The text spans in this work are claim spans in patent claims. A patent claim defines the scope of the legal protection conferred by a patent. Most of the time a patent has several claims to define its scope. The reason why it might be possible to build a function to evaluate the utility requirement is that a granted claim is presumed to have met the utility requirement. It could be said that granted patents are human-annotated and possible for supervised learning. The problem lies in how to identify and make use of such annotations.

We identify two types of human annotation in patent claims: explicit and implicit. The explicit annotation is manifested by the dependency between an independent claim and a dependent claim. For example, a dependent claim such as "2. The method of claim 1, wherein…." defines a dependency between claim 2 and claim 1. The implicit annotation is based on the property of element combination. In patent claim language, an invention could be decomposed into inventive elements and conceptually, for a classification task, the order of the elements describing how they work collectively does not matter. The identification and boundary of an inventive element is, therefore, an implicit annotation. Such a property of being able to combine elements in different order is pretty unique, compared with other mainstream NLP research.

Leveraging both the explicit and implicit of annotations is the reason why supervised learning might be feasible for learning the utility requirement. Before building training datasets, a technical problem is how to identify the inventive elements in a patent claim. The format of patent claims provides an answer. A patent claim is required to be a single sentence. Since it defines a technical scope to be protected, a patent claim is usually much longer than an ordinary sentence. Such an unusual length is a challenge to most inventors and even to patent practitioners. Therefore, it is common to split a claim into text spans. A claim span is a segment of claim text. For example, the claim 1 of US9229634B2 is divided into spans as **Fig. 1**. A claim span for readability is a suitable approximation of an inventive element. We assume such approximation sufficient for proof of concept in this work and leave finer approximation to the future, such as training a neural network to split a longer span into shorter ones.

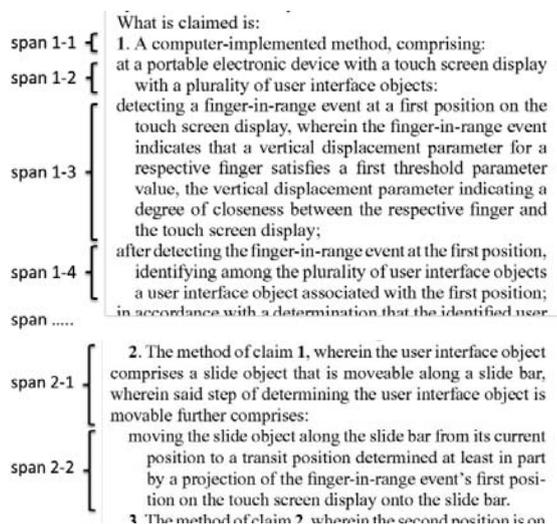

**Fig. 1.** Spans in the '634 patent

### 1.4 Span-pair Classification

Based on the utility requirement, we treat two spans in a patent claim relevant to a useful function of some kind and relevant to each other. We further take such a relevance problem as a classification problem and the classification is binary: relevant or irrelevant. Our goal is to train a neural network to predict the relevancy between two claim spans. For example, in **Fig. 1**, the span 1-1, 1-2, 1-3 and 1-4 are relevant to each other. The span 2-1 and 2-2 are relevant to each other. In addition, since the claim 2 is dependent on the claim 1, the span 2-1 and 2-2 are relevant to all spans in the claim 1 too. We collect such intra-claim span pairs and inter-claim span pairs and build a dataset of relevant span pairs.

Training a binary classifier needs both positive records and negative records. The relevant span pairs are positive records. As for negative records, we leverage the current patent classification system, such as CPC (Cooperative Patent Classification),[1] and apply negative sampling to select claim spans from non-overlapped patent classes. For example, the subclass labels of the '634 patent are G06F and H04M. The claim spans from patents without any of the same labels are sampled randomly. We assume that such claim spans would be negative records and suitable for building a dataset of irrelevant span pairs. After having both of the relevant or irrelevant span pairs, we formulate the span-pair relevancy problem as a sentence-pair classification problem.

It is noted that patents with different subject matters, such as process, machine, manufacture, and composition of matter, may have different average span lengths or different average numbers of spans per claim. We avoid any manual feature engineering and hypothesize that the neural network can learn different subject matters in one model well based on two assumptions: (1) The span boundaries are annotated by humans for easier comprehension. Therefore, it should be feasible for the neural network to learn such comprehensible spans. It might also be possible that the neural network can estimate what kind of subject matter a patent claim is. (2) As mentioned in Section 1.3, it should be possible to train a neural network to calculate more fine-grained span boundaries. By doing so, the relation between spans in different subject matters should be similar from a data perspective. In our previous works [1] [2] and this work, we did not encounter any issue requiring different treatments for different subject matters. We leave the validation of the aforementioned assumptions to the future.

### 1.5 Patent Claim Generation with GPT-2

Initially, we tried patent claim generation at span level by using the span-pair classifier. We thought that, by ranking the relevancy with respect to all existing spans, it might be possible to provide candidate spans based on user's input. Unfortunately, it did not work. As a binary classifier, its relevancy ranking is polarized to binary results most of the time. The neural network cannot produce ranking results with finer granu-larity. We switched gear to the OpenAI GPT-2 by Radford et al. [3] for text generation word by word. In order to use the span-pair classifier to check the generated patent claims by GPT-2, we add special span separators to patent claims. In this way, GPT-2 can generate patent claims with span separators which make it possible to split a generated patent claim into claim spans. We use the span-pair classifier to check whether two consecutive spans of the generated patent claim are relevant or not.

---

[1] https://www.cooperativepatentclassification.org

For details in patent claim generation, please refer to our previous work [1]. In brief, our patent claim generation is based on the unconditional random sampling in GPT-2. The sampling is a *top_k* random sampling algorithm with a default *k* value 40. It means sorting by probability and zeroing out anything below the 40th token when sampling. The quality of generated text depends on the distribution of reasonable words in the top 40 tokens. If there are many words one could sample from reasonably, the quality would be higher. If there are only a few reasonable words to sample from, the quality would be lower. We experiment with different *k* values in this work. We also hypothesize that a higher *k* value will generate patent claims with higher diversity and the span-pair relevancy will, therefore, become lower. To our knowledge, our previous work [1] is the first to generate patent claims by transfer learning with Transformer [4] models. This work is the first to observe the correlation between the diversity of text generation by one Transformer and the span relevancy measured by the other Transformer model.

## 2    Framework

We propose a generic framework for text generation and quality measurement based on Transformer architecture. In **Fig. 2**, on the right-hand side, the quality measurement is based on a fine-tuned Transformer Encoder. The Encoder is the BERT model by Devlin et al. [5]. BERT is a language representation model which stands for Bidirectional Encoder Representations from Transformers. We build the fine-tuned BERT by fine-tuning the pre-trained BERT model released by Google.[2] In Section 3, we describe how to build a span-pair classifier based on BERT. On the left-hand side, the text generation is based on a fine-tuned Transformer Decoder. The Decoder is the GPT-2 model. We build the fine-tuned GPT-2 by fine-tuning the pre-trained GPT-2 model released by OpenAI.[3] In Section 4, we describe how to work on GPT-2 in more details. By using different *k* values for the *top_k* function in GPT-2, we can generate patent claims with different randomness. The next step is to split the generated patent claims into span pairs. The span pairs are then fed to the fine-tuned BERT for measuring relevancy.

In this framework, we treat the fine-tuned models as building blocks and replaceable. In fact, it is preferable but not necessary to use Transformer Decoder for text generation and use Transformer Encoder for quality measurement. A Transformer Encoder can be used for text generation. A Transformer Decoder can be used for quality measurement too. If the Encoder and the Decoder of the same Transformer are

---
[2] https://github.com/google-research/bert
[3] https://github.com/openai/gpt-2

co-trained, it is also possible to use the same Transformer for both text generation and quality measurement. Whether this kind of replacement is better is another research topic. At the moment of this writing, BERT and GPT-2 are the best available Encoder and Decoder, respectively. For example, using BERT for text generation is feasible by Wang and Cho [6], but it does not outperform GPT-2. There are several competing

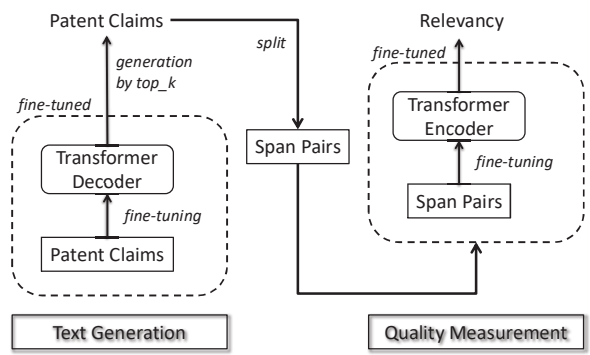

**Fig. 2.** Framework of Text Generation & Measurement

Transformer-based models emerging too, such as RoBERTa [7], MASS [8], XLNet [9] and ERNIE 2.0 [10]. This is the reason why we generalize our implementation from the perspective of a framework. Our current work is a baseline for benchmarking in the future.

## 3   Build a Span-pair Classifier based on BERT

Sentence pair classification has been a type of problem for a long time. Conventionally the goal of the sentence pair classification is to predict whether the second sentence is an entailment, contradiction, or neutral. In this work, we simplify the prediction as a binary classification to entail relevancy or irrelevancy in terms of the utility requirement for patents. One recent approach to text classification problems is fine-tuning a pre-trained BERT model. Such an approach produced the state-of-the-art models for several sentence pair classification tasks, e.g. MNLI, QQP, QNLI, SST-2, CoLA, STS-B, MRPC, and RTE. Therefore, we leverage this fine-tuning approach to build our span-pair classifier. Besides, such a fine-tuning approach applies to other text classification problems as well. For example, a new state-of-the-art result for patent classification based on BERT is produced in our previous work [2].

### 3.1 Data Pipeline

A data pipeline of preprocessing is required in order to split the raw text of patent claims into span pairs. There are three stages in our data pipeline: (1) raw data collection, (2) claim span identification, and (3) span combination. Although raw data is available on the USPTO Open Data Portal,[4] we found it easier to leverage the Google Patents Public Datasets on BigQuery.[5] A dataset based on SQL lowers the entry barrier of data preparation. In the appendix, we share our SQL statements as a better way than sharing conventional raw data for two reasons: (1) Separation of concerns. If the shared raw data contains pre-processing or post-processing specific to a problem or solution, it will be harder for other researchers to reuse for different data processing. (2) Clarity and flexibility. A SQL statement is more precise and easier to revise.

At the second stage, we split patent claim text into claim spans on a heuristic basis. Very often, patent practitioners use semicolon or comma to separate a long patent claim into multiple lines. The character return between lines was omitted in our queried data. Therefore, a convenient trick is to make use of such omissions to identify claim spans. Such a heuristic span identification might be not perfect, but this kind of data approximation is sufficient for proof of concept in our work.

At the third stage, the span combination includes two types of data preparation: positive sampling and negative sampling. Generating relevant span pairs as positive records is more intuitive. If the claim is an independent claim having *n1* spans, *n1\*(n1-1)* span pairs will be generated. If the claim is a dependent claim having *n2* spans and depending on a claim having *n3* spans, *n2\*n3* spans will be generated in addition to *n2\*(n2-1)* spans. We skipped the scenario of multiple dependencies since it does not occur often.

Generating irrelevant span pairs (span1, span2) as negative records is trickier. The number of records based on negative sampling is twice the number of records of positive sampling because we define two types of negative sampling. Type (a) is to randomly pick a span2 which does not share any CPC Subclass label (e.g., "G06F") with the span1. Under the CPC hierarchy, a Section label is higher and more inclusive than a Subclass label. Based on this hierarchy, we design type (b) to pick another span2 which shares the same Section label (e.g., "G") with the span1 but does not share any Subclass label. We hypothesize that the two spans in type (a) could be farther away in tensor space and easier for a neural network to learn. By adding the type (b), the two spans could be closer but still irrelevant. This may make the neural network generalize better. We leave the validation of this hypothesis to the future.

One more consideration at the span combination stage is to avoid data explosion. In fact, a patent may have many claims and a claim may have many spans. In this work, we cap both the number of claims per patent and the number of spans per claim to 20. In addition, due to resource constraint, our datasets cover the first month of year 2013 to 2016 only. The number of span pairs in our datasets are 3,826,027 (2013), 3,257,617 (2014), 3,316,532 (2015) and 3,728,207 (2016) respectively, as shown in **Table 1**. We found such datasets sufficient to show the stability of accuracy in our results. We leave broader coverages and different approaches of span combination to the future.

---
[4] https://developer.uspto.gov/
[5] https://console.cloud.google.com/bigquery?p=patents-public-data

### 3.2 Method & Experimental Setup

According to Devlin et al. [5], a pre-trained BERT model can be fine-tuned with just one additional output layer to create state-of-the-art models for a wide range of tasks, such as question answering and language inference, without substantial task-specific architecture modifications. The fine-tuning approach introduces minimal task-specific parameters and is trained on the downstream tasks by simply fine-tuning all pre-trained parameters. Computer vision research has also demonstrated the importance of transfer learning from large pre-trained models, where an effective recipe is to fine-tune pre-trained models.

In this work, we follow the transfer learning technique and leverage the released BERT-Base pre-trained model (Uncased: 12-layer, 768-hidden, 12-heads, 110M parameters). We leave other bigger models such as the BERT-Large (340M parameters) to the future since the BERT-Base is sufficient for proof of concept. Our implementation follows the same fine-tuning example and source code released in the BERT project. In order to build a vanilla baseline for future experiments to compare with, we reduce the number of training epoch from 3 to 1 and change the max_seq_length value from 128 to 256 to reflect the fact that claim spans are longer. All other hyperparameters remain the same.

### 3.3 Results

We show the performance of the span pair classifier in **Table 1**. Each dataset contains both positive sampling (relevant span pairs) and negative sampling (irrelevant span pairs). The accuracy of evaluating relevant span pairs stays above 84% at least. The accuracy of evaluating both relevant and irrelevant span pairs stays above 92%. Such numbers are significant. Take row 3 as an example. The fine-tuning dataset belongs to year 2013 and, at each column, the testing datasets belong to year 2013, 2014, 2015 and 2016 respectively. After fine-tuned, the testing accuracy of dataset 2014 is 93.07% if covering positive and negative samplings. The testing accuracy is 87.19% if covering positive sampling only. The testing accuracies of dataset 2015 are 92.79% and 86.37% respectively. The testing accuracies of dataset 2016 are 92.74% and 86.24% respectively. We think that such accuracy is sufficient to measure text generation in terms of span relevancy.

Table 1. Evaluation Result (accuracy in percentage)

| Test | 2013 | | 2014 | | 2015 | | 2016 | |
|---|---|---|---|---|---|---|---|---|
| Fine-tuning | pos & neg | pos only | pos & neg | pos only | pos & neg | pos only | pos & neg | pos only |
| 2013 | 97.09 | 96.08 | 93.07 | 87.19 | 92.79 | 86.37 | 92.74 | 86.24 |

| 2014 |  |  | 97.09 | 95.73 | 92.56 | 84.99 | 92.44 | 84.87 |
| 2015 |  |  |  |  | 96.99 | 95.90 | 92.70 | 85.94 |

- pos = positive sampling (i.e., relevant span pairs),
- neg = negative sampling (i.e., irrelevant span pairs)
- Dataset is limited to January, due to cloud resource constraint. Number of records:
  - 2013: 3,826,027 (pos & neg), 1,412,983 (pos only)
  - 2014: 3,257,617 (pos & neg), 1,128,790 (pos only)
  - 2015: 3,316,532 (pos & neg), 1,158,265 (pos only)
  - 2016: 3,728,207 (pos & neg), 1,364,092 (pos only)

## 4 Measuring Text Generation by Span Relevancy

Before measuring text generation, we have to build a fine-tuned GPT-2 model for text generation first. We leverage the same codebase in our previous work [1] and fine-tune the pre-trained 345M model released by OpenAI again. In the followings, we describe the dataset for fine-tuning, parameters to experiments and our measurement results.

### 4.1 Data

Our dataset to fine-tune the pre-trained GPT-2 model contains 18,000 patent claims. It is composed of three smaller datasets, and each dataset has 6,000 records. The first smaller dataset belongs to CPC Class A (Human Necessities) and its data period is from year 2011 to 2014. The second one belongs to CPC Class G (Physics) and has the same data period. The third one belongs to both CPC Class A and G. Due to fewer records being available, the data period for the third dataset is from 2000 to 2016 so that 6,000 records can be collected. These three datasets are prepared for our next paper in which we plan to observe the detail transition in GPT-2 fine-tuning. For possible integration of these two papers in the future, we decide to leverage the same datasets in this work.

### 4.2 Method & Experimental Setup

According to Radford et al. [3], GPT-2 is a successor to GPT (Generative Pre-Training) which uses a Transformer-based architecture as its language model. GPT-2 largely follows the details of the original GPT model with a few modifications, particularly a larger model size and more training data. When a large language model is trained on a sufficiently large and diverse dataset, it is able to perform well across many domains and datasets. In GPT-2, text generation is based on *top_k* random sampling. The random sampling can be either conditional or unconditional. The authors use *k* value as 40 in their experiments and show several state-of-the-art results.

In our work, we tested the *top_k* unconditional sampling in GPT-2 with different *k* values: 3, 40, 100, 1000 and 10000. When fine-tuning the pre-trained GPT-2 model, we keep most of the hyperparameters in our previous work [1] and we set warmup step as 1,000 and training step as 10,000 to reach a reasonable training loss empirically. After fine-tuned, for each experiment, we generate 512 patent claims. These patent claims contain span separators since the GPT-2 model has been fine-tuned with patent claims containing span separators. Based on the output, we split a generated patent claim into spans and combine two consecutive claim spans as a span pair. On average, each patent claim contains 4.34 span pairs.

### 4.3 Results

The purpose of this work is to propose a generic framework for text generation and quality measurement. Measuring the overall quality of text generation is a challenging problem. In this work, we treat span relevancy as a global quality metric and also a proof of concept of the framework. It is open for future researchers to devise more models and metrics to measure more qualities of text generation. Based on GPT-2 and BERT, **Table 2** shows our experiment results. The relevancy ratio in the table is defined as the total number of relevant pairs divided by the total number of all generated pairs. The results validate our hypothesis that a higher k value (more randomness) for text generation will produce a lower relevancy ratio between spans. For example, the ratio is 92.67% (highest) when k is 3 (lowest), and the ratio is 76.97% (lowest) when k is 10,000 (highest). Between these two experiments, the relevancy ratio decreases when the randomness in GPT-2 increases. Such a correlation between randomness and relevancy is intuitive, and it is probably observable if mechanical Turk is involved for manual evaluations. However, based on human's subjective and qualitative reviews, it should be hard to have an objective and quantitative value to show the difference in text generation quality. To our knowledge, this work is the first to define a quality metric at span level for GPT-2.

**Table 2.** Relevancy Ratio of Generated Patent Claims

| *top_k* | 3 | 40 | 100 | 1,000 | 10,000 |
|---|---|---|---|---|---|
| **relevant pairs** | 2,592 | 2,117 | 1,992 | 1,671 | 1,401 |
| **irrelevant pairs** | 205 | 202 | 210 | 323 | 419 |
| **relevancy ratio** | 92.67% | 91.28% | 90.83% | 83.80% | 76.97% |

According to OpenAI, it is noteworthy that GPT-2 can generate pretty coherent text which outperforms prior techniques in the past. The default *k* value in the original GPT-2 codebase is 40. In our experiment, the relevancy ratio for *k=40* is 91.28%. Such a high ratio aligns with the public perception of GPT-2's coherent text generation. For patent claim generation, it can be further noted that a higher relevancy ratio might be not necessarily always better. The ratio reflects a tradeoff between diversity and coherency. The preferred range of relevancy ratio should depend on downstream tasks. For example, if the relevancy ratio is too high, what generated might be similar to each other and within a limited range in terms of creativity in patent claims. If the relevancy ratio is too low, what generated might make no practical sense at all. We leave it an open question regarding what range of relevancy ratio should be suitable for better patent claim generation.

## 5 Conclusion

Patents might be an ideal data source for inventors to move toward human-machine co-inventing in the long run. The emergence of Transformer models such as BERT and GPT-2 is a paradigm shift and a tremendous opportunity for patent researchers. Our contributions in this work include: (1) proposing a framework of using one Transformer to measure the other Transformer, (2) fine-tuning a pre-trained BERT model as a classifier of span-pair relevancy, and (3) using the classifier to measure the patent claims generated by a pre-trained GPT-2 model. Our result validates the quantitative metric of relevancy in patent claim generation. Notably, the span relevancy ratio calculated by BERT becomes higher when the diversity in GPT-2 text generation becomes lower. By having a way to measure text generation quantitatively, we expect to push text generation quality further in the future.

## 6 Appendix

The following SQL selects the first claims of all US utility patents in 2013 and aggregates the CPC codes at subclass level: (source: Google Patents Public Datasets[6])

```
SELECT STRING_AGG(distinct t2. group_id order by t2. group_id)
AS cpc_ids, t1.id, t1.date, text
  FROM `patents-public-data.patentsview.patent` t1,
  `patents-public-data.patentsview.cpc_current` t2,
  `patents-public-data.patentsview.claim` t3
  where t1.id = t2.patent_id and t1.id = t3.patent_id
  and timestamp(t1.date) >= timestamp('2013-01-01')
  and timestamp(t1.date) <= timestamp('2013-12-31')
  and t3.sequence='1' and t1.type='utility'
  group by t1.id, t1.date, t3.text
```

---

[6] https://console.cloud.google.com/bigquery?p=patents-public-data

## Acknowledgements

We thank the anonymous reviewers for their valuable feedback. The research reported in this manuscript has been funded by the Ministry of Science and Technology (MOST) in Taiwan (Project: 108-2221-E-002-104-MY3).

## References

1. Lee, J.-S., Hsiang, J.: Patent Claim Generation by Fine-Tuning OpenAI GPT-2. (2019).
2. Lee, J.-S., Hsiang, J.: PatentBERT: Patent Classification with Fine-Tuning a pre-trained BERT Model. (2019).
3. Radrof, A., Wu, J., Child, R., Luan, D., Amodei, D., Sutskever, I.: Language Models are Unsupervised Multitask Learners. (2018).
4. Vaswani, A., Shazeer, N., Parmar, N., Uszkoreit, J., Jones, L., Gomez, A.N., Kaiser, Ł., Polosukhin, I.: Attention is all you need. Adv. Neural Inf. Process. Syst. 2017-Decem, 5999–6009 (2017).
5. Devlin, J., Chang, M.-W., Lee, K., Toutanova, K.: BERT: Pre-training of Deep Bidirectional Transformers for Language Understanding. In: Proceedings of the 2019 Conference of the North American Chapter of the Association for Computational Linguistics: Human Language Technologies, NAACL-HLT 2019, Minneapolis, MN, USA, June 2-7, 2019, Volume 1 (Long and Short Papers). pp. 4171–4186 (2019).
6. Wang, A., Cho, K.: BERT has a Mouth, and It Must Speak: BERT as a Markov Random Field Language Model. In: Proceedings of the Workshop on Methods for Optimizing and Evaluating Neural Language Generation. pp. 30–36. Association for Computational Linguistics, Minneapolis, Minnesota (2019). https://doi.org/10.18653/v1/W19-2304.
7. Liu, Y., Ott, M., Goyal, N., Du, J., Joshi, M., Chen, D., Levy, O., Lewis, M., Zettlemoyer, L., Stoyanov, V.: RoBERTa: A Robustly Optimized BERT Pretraining Approach. (2019).
8. Song, K., Tan, X., Qin, T., Lu, J., Liu, T.-Y.: MASS: Masked Sequence to Sequence Pre-training for Language Generation. In: Proceedings of the 36 th International Conference on Machine Learning (2019).
9. Yang, Z., Dai, Z., Yang, Y., Carbonell, J., Salakhutdinov, R., Le, Q.V.: XLNet: Generalized Autoregressive Pretraining for Language Understanding. (2019).
10. Sun, Y., Wang, S., Li, Y., Feng, S., Tian, H., Wu, H., Wang, H.: ERNIE 2.0: A Continual Pre-training Framework for Language Understanding. (2019).